%% file: icde.tex
\documentclass[conference]{IEEEtran}
\IEEEoverridecommandlockouts
\usepackage[numbers]{natbib}
\usepackage{url}
\usepackage{amsmath}
\usepackage{mathtools}
\usepackage{filecontents}
\usepackage{amsmath,amssymb,amsfonts}
\usepackage{algorithmic}
\usepackage{graphicx}
\usepackage{multirow}
\usepackage{subfig}
\usepackage{xcolor}
\usepackage{textcomp}
\usepackage{microtype}
\usepackage[ruled,linesnumbered,noend]{algorithm2e}

\newtheorem{definition}{Definition}
\usepackage{tabularx}
\usepackage{adjustbox}
\usepackage{xspace,colortbl}
\usepackage{amssymb}
\usepackage{amsmath}
\usepackage{mathtools}
\usepackage{adjustbox}

\usepackage{bm}
\usepackage{multirow}
\usepackage[ruled,linesnumbered,noend]{algorithm2e}
\usepackage{array}
\usepackage{subfig}
\usepackage{graphicx}
\usepackage{dblfloatfix}

\usepackage{float}

\makeatletter
\def\footnoterule{\kern-3\p@
  \hrule \@width 2in \kern 2.6\p@} 
\makeatother

\def\BibTeX{{\rm B\kern-.05em{\sc i\kern-.025em b}\kern-.08em
    T\kern-.1667em\lower.7ex\hbox{E}\kern-.125emX}}
\begin{document}

\title{Temporal Network Representation Learning via Historical Neighborhoods Aggregation}

\author{\IEEEauthorblockA{Shixun Huang\textsuperscript{$\dagger$} ~~ Zhifeng Bao\textsuperscript{$\dagger$} ~~ Guoliang Li\textsuperscript{$\ddagger$}~~ Yanghao Zhou\textsuperscript{$\dagger$} ~~ J. Shane Culpepper\textsuperscript{$\dagger$} }
\IEEEauthorblockA{\textsuperscript{$\dagger$}\textit{RMIT University, Australia} ~~~~
\textsuperscript{$\ddagger$}\textit{Tsinghua University, China}\\
}
}

\maketitle
\input{./tex/macros.tex}

\input{./tex/abstract.tex}

\input{./tex/introduction.tex}

\input{./tex/relatedwork.tex}

\input{./tex/preliminary.tex}

\input{./tex/method.tex}

\input{./tex/experiment.tex}

\bibliographystyle{IEEEtran}
\bibliography{library}

\end{document}

%% file: tex/macros.tex
\def\D{\hphantom{1}}
\def\C{\hphantom{1,}}
\newcommand\method[1]{{\textbf{\sf{\sc #1}}}}
\newcommand\cmethod[1]{{\textbf{\sf{\sc #1}}}}
\newcommand\smethod[1]{{\textbf{\sf\scriptsize{#1}}}}
\newcommand{\define}[1]{\textbf{\textsl{#1}}}
\newcommand{\guoliang}[1]{\textrm{\textcolor{green}{Guoliang says: #1}}}
\newcommand{\bao}[1]{\textrm{\textcolor{red}{Bao says: #1}}}
\newcommand{\huang}[1]{\textrm{\textcolor{blue}{ #1}}}
\newcommand{\bit}[1]{\mathfrak{B}^#1}
\newcommand{\cb}{containment bitset }
\newcommand{\cbs}{containment bitsets }
\newcommand{\lb}{local containment bitset }
\newcommand{\lbs}{local containment bitsets }
\newcommand{\tb}{traversal bitset }
\newcommand{\tbs}{traversal bitsets }
\newcommand{\rb}{recording bitset }
\newcommand{\myparagraph}[1]{\vspace{0.3\baselineskip}\noindent{\textbf{#1.}}~}
\newcommand{\var}[1]{\mbox{\emph{#1}}}
\newcommand{\svar}[1]{\mbox{\scriptsize\emph{#1}}}
\newcommand{\tvar}[1]{\mbox{\tiny\emph{#1}}}
\newcommand{\avar}[1]{\mbox{#1}}
\newcommand{\asvar}[1]{\mbox{\scriptsize{#1}}}
\newcommand{\atvar}[1]{\mbox{\tiny{#1}}}

%% file: tex/abstract.tex
\begin{abstract}
Network embedding is an effective method to learn low-dimensional
representations of nodes, which can be applied to various real-life
applications such as visualization, node classification, and link
prediction.
Although significant progress has been made on this problem in recent
years, several important challenges remain, such as how to properly
capture temporal information in evolving networks.
In practice, most networks are continually
evolving.
Some networks only add new edges or nodes such as authorship
networks, while others support removal of nodes or edges such as
internet data routing.
If patterns exist in the changes of the network structure,
we can better understand the relationships between nodes and the
evolution of the network, which can be further leveraged to learn node
representations with more meaningful information.
In this paper, we propose the Embedding via Historical Neighborhoods
Aggregation (EHNA) algorithm.
More specifically, we first propose a temporal random walk that can
identify relevant nodes in historical neighborhoods which have impact
on edge formations.
Then we apply a deep learning model which uses a custom attention
mechanism to induce node embeddings that directly capture temporal
information in the underlying feature representation.
We perform extensive experiments on a range of real-world datasets, and the
results demonstrate the effectiveness of our new approach in the 
network reconstruction task and the link prediction task.

\setcounter{footnote}{1}
\footnotetext{Zhifeng Bao is the corresponding author.}
\end{abstract}

%% file: tex/introduction.tex
\section{introduction}
Network embedding has become a valuable tool for solving a wide
variety of network algorithmic problems in recent years.
The key idea is to learn low-dimensional representations for nodes in
a network.
It has been applied in various applications, such as link prediction
\cite{grover2016node2vec}, network reconstruction
\cite{wang2016structural}, node classification
\cite{perozzi2014deepwalk} and visualization~\cite{tang2015line}.
Existing studies
\cite{grover2016node2vec,wang2016structural,perozzi2014deepwalk,tang2015line}
generally learn low-dimensional representations of nodes over a
static network structure by making use of contextual information
such as graph proximity.

However, many real-world networks, such as co-author networks and
social networks, have a wealth of temporal information (e.g., when
edges were formed).
Such temporal information provides important insights into the
dynamics of networks, and can be combined with contextual information
in order to learn more effective and meaningful representations of
nodes.
Moreover, many application domains are heavily reliant on temporal
graphs, such as instant messaging networks and financial transaction
graphs, where timing information is critical.
The temporal/dynamic nature of social networks has attracted
considerable research attention due to its importance in many problems, including  
dynamic personalized pagerank~\cite{guo2017parallel},
advertisement recommendation~\cite{li2016context} and temporal
influence maximization~\cite{huang2019finding}, to name a few.
Despite the clear value of such information, temporal network
embeddings are rarely applied to solve many other important tasks
such as network reconstruction and link
prediction~\cite{hamilton2017representation,cui2018}.

\begin{figure}[!t]
\centering
\includegraphics[width=0.4\textwidth]{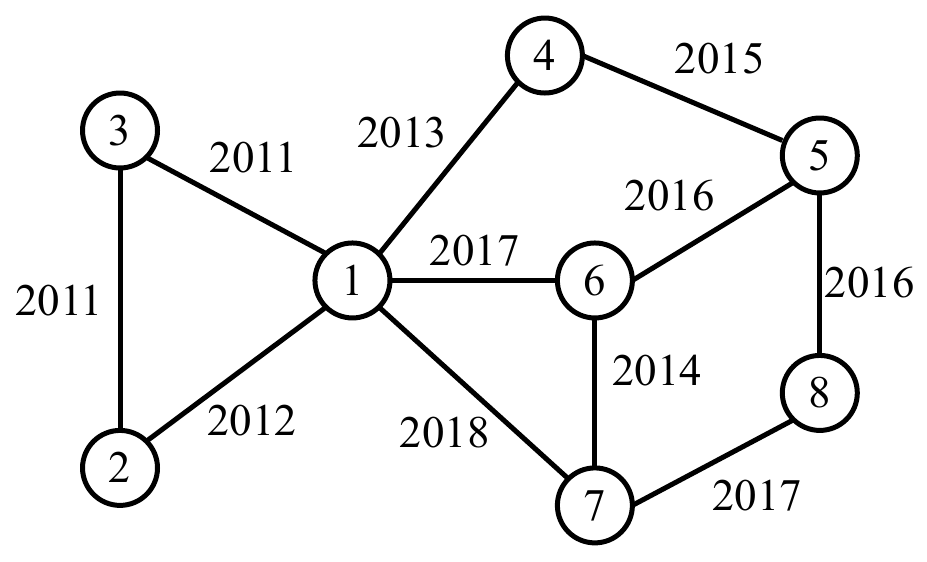}
\caption{An example of a co-author temporal network.
Each edge is annotated with a timestamp denoting when the edge was
created.
\label{coauthor}} 
\end{figure}

\begin{figure}[!t]
\centering
\includegraphics[width=0.4\textwidth]{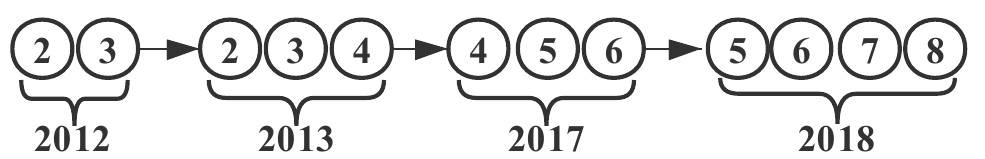}
\caption{The evolution of a set of nodes that are contextually and
temporally related to node 1.}\label{neighbor}
\end{figure}

Incorporating temporal information into network embeddings can improve
the effectiveness in capturing relationships between nodes which can
produce performance improvements in downstream tasks.
To better understand the relationships between nodes with temporal
information in a network, consider Figure~\ref{coauthor} which is an ego co-author network for a node (1).
Each node is an author, and edges represent collaborations denoted
with timestamps.
Without temporal knowledge, the relationships between nodes 2, 3, 4,
6 and 7 are indistinguishable since they are all connected to node 1.
However, when viewed from the temporal perspective, node 1 once was `close' to nodes 2
and 3 but now is `closer' to nodes 4, 6 and 7 as node 1 has more
recent and frequent collaborations with the latter nodes.
Furthermore, in the static case, nodes 2 and 3 appear to be `closer'
to node 1 than node 5, since nodes 2 and 3 are direct neighbors of
node 1, whereas node 5 is not directly connected to node 1.
A temporal interpretation suggests that node 5 is also important to 1
because node 5 could be enabling collaborations between nodes 1 6,
and 7.
Thus, with temporal information, new interpretations of the
`closeness' of relationships between node 1 and other nodes and how
these relationships evolve are now possible.

To further develop this intuition, consider the following concrete
example.
Suppose node 1 began publishing as a Ph.D.\ student.
Thus, papers produced during their candidature were co-authored with
a supervisor (node 3) and one of the supervisor's collaborators (node
2).
After graduation, the student (node 1) became a research scientist in
a new organization.
As a result, a new collaboration with a senior supervisor (node 4) is
formed.
Node 1 was introduced to node 5 through collaborations between the
senior supervisor (node 4) and node 5, and later began working with
node 6 after the relationship with node 5 was formed.
Similarly, after a collaboration between node 1 and node 6, node 1
met node 7 because of the relationships between node 5 and node 8,
node 8 and node 7, and node 6 and node 7.
Hence, as shown in Figure~\ref{neighbor}, both the formation of the
relationships between node 1 and other nodes and the levels of
closeness of such relationships evolve as new connections are made
between node 1 and related collaborators.

Thus, the formation of each edge $(x,y)$ changes not only the
relationship between $x$ and $y$ but also relationships between nodes
in the surrounding neighborhood.
Capturing how and why a network evolves (through edge formations) can
produce better signals in learned models.
Recent studies capture the signal by periodically probing a network
to learn more meaningful embeddings.
These methods model changes in the network by segmenting updates into
fixed time windows, and then learn an embedding for each network
snapshot~\cite{zhou2018dynamic,zhu2016scalable}.
The embeddings of previous snapshots can be used to infer or indicate
various patterns as the graph changes between snapshots.
In particular, \method{CTDNE}~\cite{nguyen2018continuous} leveraged
temporal information to sample time-constrained random walks and
trained a skip-gram model such that nodes co-occurring in these
random walks produce similar embeddings.
Inspired by the Hawkes process~\cite{hawkes1971spectra} which shows
that the occurrence of current events are influenced by the historical
events, \method{HTNE}~\cite{zuo2018embedding} leveraged prior
neighbor formation to predict future neighbor formations.
In this paper, we show how to more effectively embed fine-grained
temporal information in {\em all} learned feature representations
that are directly influenced by historical changes in a network.

How to effectively analyze edge formations is a challenging problem,
which requires: (1) identifying relevant nodes (not limited to direct
neighbors) which may impact edge formations; (2) differentiating the
impact of relevant nodes on edge formations; (3) effectively
aggregating useful information from relevant nodes given a large
number of useful but noisy signals from the graph data.

To mitigate these issues, we propose a deep learning method, which we refer
to as Embedding via Historical Neighborhoods Aggregation
(\method{EHNA}), to capture node evolution by probing the
information/factors influencing temporal behavior (i.e., edge
formations between nodes).
To analyze the formation of each edge $(x,y)$, we apply the the
following techniques.
First, a temporal random walk is used to dynamically identify
relevant nodes from historical neighborhoods that may influence new
edge formations.
Specifically, our temporal random walk algorithm models the relevance
of nodes using a configurable time decay, where nodes with high
contextual and temporal relevance are visited with higher
probabilities.
Our temporal random walk algorithm can preserve the breadth-first
search (BFS) and depth-first search (DFS) characteristics of
traditional solutions, which can then be further exploited to provide
both micro and macro level views of neighborhood structure for 
the targeted application.
Second, we also introduce a two-level (node level and walk
level) aggregation strategy combined with stacked LSTMs to
effectively extract features from nodes related to chronological
events (when edges are formed).
Third, we propose a temporal attention mechanism to improve the
quality of the temporal feature representations being learned based
on both contextual and temporal relevance of the nodes.

Our main contributions to solve the problem of temporal network
representation learning are:
\begin{itemize}
\item We leverage temporal information to analyze edge formations
such that the learned embeddings can preserve both structural network
information (e.g., the first and second order proximity) as well as
network evolution.
\item We present a temporal random walk algorithm which dynamically
captures node relationships from historical graph neighborhoods.
\item We deploy our temporal random walk algorithm in a stacked LSTM
architecture that is combined with a two-level temporal attention and
aggregation strategy developed specifically for graph data, and
describe how to directly tune the temporal effects captured in
feature embeddings learned by the model.
\item We show how to effectively aggregate the resulting temporal
node features into a fixed-sized readout layer (feature embedding),
which can be directly applied to several important graph problems
such as link prediction.
\item We validate the effectiveness of our approach using four
real-world datasets for the network reconstruction and link
prediction tasks.
\end{itemize}

\smallskip
\noindent \textbf{Organization of this paper.}
Related work is surveyed in Section~\ref{relatedwork} and our problem
formulation is presented in Section~\ref{preliminary}.
Next, we describe the problem solution in
Section~\ref{proposedmethod}.
We empirically validate our new approach in Section~\ref{experiment},
and conclude in Section~\ref{conclusion}.

%% file: tex/relatedwork.tex
\section{Related Work}\label{relatedwork}
\textbf{Static network embedding.}
Inspired by classic techniques for dimensionality reduction
\cite{belkin2002laplacian} and multi-dimensional
scaling~\cite{kruskal1964multidimensional}, early methods
\cite{belkin2002laplacian,ahmed2013distributed,cao2015grarep,ou2016asymmetric}
focused on matrix-factorization to learn graph embeddings.
In these approaches, node embeddings were learned using a
deterministic measure of graph proximity which can be approximated by
computing the dot product between learned embedding vectors.
Methods such as \method{DeepWalk}~\cite{perozzi2014deepwalk} and
\method{Node2Vec}~\cite{grover2016node2vec} employed a more flexible,
stochastic measure of graph proximity to learn node embeddings such
that nodes co-occurring in short random walks over a graph produce
similar embeddings.
A more contemporaneous method, \method{Line}~\cite{tang2015line},
combined two objectives that optimize `first-order' and
`second-order' graph proximity, respectively.
\cmethod{sdne}~\cite{wang2016structural} extended this idea of
\method{Line} to a deep learning framework~\cite{hinton2006reducing}
which simultaneously optimizes these two objectives.
A series of
methods~\cite{dong2017metapath2vec,tang2015pte,chen2018pme} followed
which further improved the effectiveness and/or scalability of this
approach.

More recently, Graph Convolutional Networks (\cmethod{gcn}) were
proposed and are often more effective than the aforementioned methods
for many common applications, albeit more computationally demanding.
These models can directly incorporate raw node features/attributes
and inductive learning into nodes absent during the initial training
process.
\cmethod{gcn} methods generate embeddings for nodes by aggregating
information from neighborhoods.
This aggregation process is called a `convolution' as it represents a
node as a function of the surrounding neighborhood, in a manner
similar to center-surrounded convolutional kernels commonly used in
computer vision~\cite{kipf2016semi,hamilton2017inductive}.
To apply convolutional operations on graph data, Bruna et
al.~\cite{bruna2013spectral} computed the point-wise product of a
graph spectrum in order to induce convolutional kernels.
To simplify the spectral method, Kips et al.~\cite{kipf2016semi}
applied a Fourier transform to graph spectral data to obtain filters
which can be used to directly perform convolutional operations on
graph data.
\method{GraphSAGE}~\cite{hamilton2017inductive} proposed a general
learning framework which allows more flexible choices of aggregation
functions and node combinations such that it gave significant gains
in some downstream tasks.
There are also several related \cmethod{gcn} based
methods~\cite{kipf2016semi,kipf2016variational,schlichtkrull2018modeling,BergKW17,pham2017column,GAT,chen2019}
which differ primarily in how the aggregation step is performed.
Despite significant prior work, all of these methods focused on
preserving static network information.

\textbf{Dynamic network embedding.} 
Several recent methods have also been proposed to model the dynamic
nature of networks.
Zhu et al.~\cite{zhu2016scalable} proposed a matrix
factorization-based algorithm for dynamic network embedding.
Yang et al.~\cite{zhou2018dynamic} explored the evolutionary patterns
of triads to capture structural information to learn latent
representation vectors for vertices at different timesteps.
These methods~\cite{zhu2016scalable,zhou2018dynamic} modeled the
dynamics by segmenting the timelines into fixed windows and the
resulting embeddings only represent individual snapshots of a
network.
Embeddings of previous snapshots can then be used to infer or
indicate various patterns of a graph as it changes between snapshots.
\cmethod{ctdne}~\cite{nguyen2018continuous} extended the well-known
Node2Vec~\cite{grover2016node2vec} embedding approach to temporal
data by imposing a temporal ordering restriction on the random walk
process such that random walks are one-directional (increasing in
time).
The skip-gram model adopted by
{\cmethod{ctdne}}~\cite{nguyen2018continuous} attempts to induce
co-occurring nodes in a constrained random walk to produce similar
embeddings.
However, these
methods~\cite{zhou2018dynamic,zhu2016scalable,nguyen2018continuous}
only leverage temporal information at a coarse level and thus
may not fully capture the evolution of a temporal network.
Inspired by the conditional intensity function of the Hawkes
process~\cite{hawkes1971spectra} which showed that the occurrence of
a current event is influenced by prior events, \method{htne}~\cite{zuo2018embedding} modeled
neighborhood formation sequences such that the current neighbor
formation is influenced with a higher ``intensity'' by more recent
historical neighbor formations.
However, edge formations are not just influenced by directly
connected neighbors, and the process described is not able to
directly benefit from the influence of surrounding nodes unless 
direct connections are formed.

%% file: tex/preliminary.tex
\section{Problem Formulation}\label{preliminary}
In this section, we proceed to formulate our problem. 

\smallskip
\begin{definition}
(\textbf{Temporal Network}).
A temporal network is a network $\mathcal{G}=(\mathcal{V},\mathcal{E})$
where $\mathcal{V}$ denotes the set of nodes, $\mathcal{E}$ denotes
the set of edges and each edge $(x,y) \in \mathcal{V}$ is annotated
with a timestamp $t_{(x,y)}$, describing when edge $(x,y)$ was
formed.
\end{definition}
\smallskip

Given a temporal network, the evolution of new connections to node
$x$ are influenced by nodes close to $x$.
Specifically, the formation of each edge $(x,y)$ is influenced by
{\em relevant nodes} which have direct or indirect historical
interactions with $x$ or $y$ before time $t_{(x,y)}$ in a {\em
neighborhood}.

\smallskip
\begin{definition}\label{relevant}
(\textbf{Relevant Node}).
For each edge $(x,y)$, a node $w$ is relevant to the formation of
$(x,y)$ if and only if $w$ can reach $x$ or $y$ through (in)direct
historical interactions (e.g., historical edges) $I=\{ a_1-a_2-a_3-
\cdots - a_i \}$ such that $\forall (a_{j},a_{j+1}) \in I,
t_{(a_{j},a_{j+1})} \leq t_{(a_{j+1},a_{j+2})} $, where $a_1 =w$,
$a_i=x$ or $y$.
\end{definition}

\smallskip
In our problem, we will use temporal random walks to identify
relevant nodes.
In particular, we aim to capture the network evolution as well as
structural information (e.g., the first-order and second-order
proximity) in the learned node embeddings.

\smallskip
\begin{definition}
(\textbf{First-order Proximity}~\cite{tang2015line}).
The first-order proximity in a network is the local pairwise
proximity between two nodes.
For each pair of nodes linked by an edge $(x,y)$, the weight on that
edge, $w_{xy}$, indicates the first-order proximity between $x$ and
$y$.
$w_{xy}>0$ in a weighted graph and $w_{xy}=1$ in an unweighted graph.
If no edge is observed between $x$ and $y$, their first-order
proximity is 0.
\end{definition}

\smallskip

\begin{definition}
(\textbf{Second-order Proximity}~\cite{tang2015line}).
The second-order proximity between a pair of nodes $(x,y)$ in a
network is the similarity between their neighborhood network
structures.
Mathematically, let $p_u= (w_{x,1} , \ldots , w_{x,|V|} )$ denote the
first-order proximity of $x$ with all the other vertices, then the
second-order proximity between $x$ and $y$ is determined by the
similarity between $p_x$ and $p_y$.
If no vertex is linked from/to both $x$ and $y$, the second-order
proximity between $x$ and $y$ is 0.
\end{definition}

\begin{figure*}[!t]
\centering

\includegraphics[width=1\textwidth]{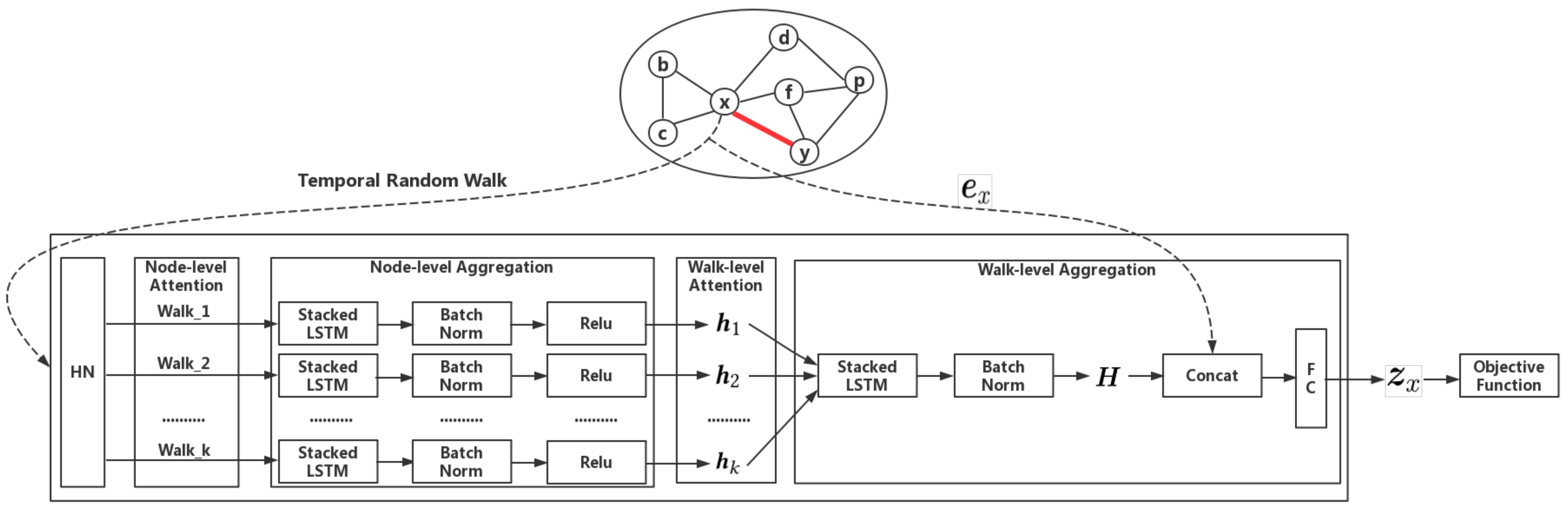}
\caption{The EHNA Framework.} \label{framework}

\end{figure*}

\smallskip
\textbf{\textsl{Problem 1 (Temporal Network Embedding)}}: Given a
temporal network $\mathcal{G}=(\mathcal{V},\mathcal{E})$, we aim to
learn a mapping function $f : v \rightarrow \bm{e}_v \in \mathbb{R}^d
$ where $v \in
\mathcal{V}, d \ll |\mathcal{V}|$.
The objective function captures:\\
(1) the Euclidean distance between $\bm{e}_x$ and
$\bm{e}_y$ to be small in the embedding space if $x$
and $y$ share an edge;\\
(2) the Euclidean distance between $\bm{e}_x$ and $\bm{e}_y$ to be
large if no link is between $x$ and $y$;\\ 
(3) $\bm{e}_x$ and $\bm{e}_y$ in the embedding space preserve the
first-order and second-order proximity of $x$ and $y$, and the
likelihood of being connected with other nodes in the future.

\smallskip

%% file: tex/method.tex
\section{Historical Neighborhood Aggregation
(\textsc{ehna})}\label{proposedmethod}
Figure~\ref{framework} provides an overview of our EHNA framework,
where HN, Concat and FC stands for the Historical Neighborhood (HN),
a concatenation operator and a fully connected layer respectively.
The {\em historical neighborhood} of node $x$ is a subgraph formed by
the nodes and edges traversed by temporal random walks initiated at
$x$.
Note that the procedure illustrated by this framework is applied to
both target nodes of every edge analyzed.
We demonstrate the main procedure using only one target node to
improve clarity.
To analyze the formation of each edge $(x,y)$, we first apply
temporal random walks to identify nodes which are relevant to the
target node $x$ ($y$).
Then we apply a two-level aggregation strategy to extract useful
temporal information that might influence future behaviors (e.g, the
formation of $(x,y)$) of $x$ ($y$).
For the first level aggregation, we apply a node-level temporal
attention mechanism to individual nodes occurring in random walks in
order to capture node-level temporal and contextual information.
During second level aggregation we induce walk-level temporal
information in the feature vectors.
Next, the \textsl{aggregated embedding} $\bm{z}_x
\;(\bm{z}_y)$ of the target node $x$ ($y$) is created by combining
the representation $\bm{e}_x \; (\bm{e}_y)$ of $x$ ($y$) with
information aggregated from the surrounding neighborhood.
Finally, we optimize our objective function using the aggregated
embeddings of the target nodes.

In the remainder of this section, we will first introduce the
temporal random walk.
Then, we explain how to achieve two-level aggregation using our
custom attention mechanism, and describe how to optimize $\bm{z}_x$ .

\subsection{Temporal Random Walk} \label{sec_randomwalk}

To analyze the formation of a target edge $(x,y)$, we use a temporal
random walk to find relevant nodes (Definition~\ref{relevant}) which
may have impacted the formation of $(x,y)$.
In our temporal random walk algorithm, the transition probability
$w^\mathcal{T}_{(u,v)}$ of each edge $(u,v)$ is directly associated
with two factors, the original weight $w_{(u,v)}$ of $(u,v)$ and the
temporal information $t_{(u,v)}$ of $(u,v)$.
To make sure that nodes that have more recent (in)direct interactions
with $x$ or $y$ are more likely to be visited in the random walks, we
use a kernel function $\mathcal{K}$ of $t_{(x,y)}$, $t_{(u,v)}$ and
$w_{(u,v)}$ to create $w^\mathcal{T}_{(u,v)}$ where
$\mathcal{K(\cdot)}$ denotes the time decay effect and can be
expressed in the form of an exponential decay function as: 

\begin{equation}
\begin{split}
w^\mathcal{T}_{(u,v)}&=\mathcal{K}(t_{(x,y)},t_{(u,v)},w_{(u,v)}) \\
                              &= w_{(u,v)} \cdot \exp(-( t_{(x,y)} -t_{(u,v)}  ))
\end{split}
\end{equation}

In a manner similar to \method{Node2Vec}~\cite{grover2016node2vec},
we introduce additional parameters to control the bias of the
temporal random walk towards different search traversals -- a
depth-first search (DFS) or a breadth-first search (BFS).
By tuning this parameter, we can ensure that the temporal random walk
captures multiple traversal paths (microscopic and macroscopic) of
the neighborhood, and the embeddings being learned can emphasize both
community and/or highly localized structural roles specific to the
target application.

For example, suppose that we require a temporal random walk which
identifies the historical neighborhood of $x$ or $y$ to analyze the
formation of $(x,y)$.
The walk has just traversed edge $(u,v)$ and now resides at node $v$.
For each outgoing edge $(v,w)$, we set the unnormalized transition
probability to $\pi_{(v,w)}= \beta_{(u,w)} \cdot
\mathcal{K}(t_{(x,y)},t_{(v,w)},w_{(v,w)})$ where 

\begin{equation}\label{random}
 \beta_{(u,w)} = \left\{ \begin{aligned}
  \frac{1}{p}&\;\;\;if \; d_{uw}=0  , \; t_{(u,v)} \geq  t_{(v,w)}\\
  1&\;\;\;if \; d_{uw}=1  , \; t_{(u,v)} \geq  t_{(v,w)}\\
  \frac{1}{q}&\;\;\;if \; d_{uw}=2  , \; t_{(u,v)} \geq  t_{(v,w)}\\
  0 &\;\;\;if \; t_{(u,v)} <  t_{(v,w)}
\end{aligned}\right.
\end{equation}

\noindent and $d_{uw}$ denotes the shortest path distance between $u$
and $w$.
By tuning the parameters $q$ and $p$, the random walk can prefer
either a BFS or DFS traversal.
More specifically, $d_{uw}$ can only have three possible values in
Equation~\ref{random}: (1) $d_{uw}=0$ refers to traveling from $v$
back to $u$; (2) $d_{uw}=1$ refers to traveling from $v$ to $w$ which
is one hop away from $u$; (3) $d_{uw}=2$ refers to traveling from $v$
to $w$ which is two hops away from $u$.
Thus, the parameter $p$ controls the likelihood of immediately
revisiting a previous node in the walk and the parameter $q$
controls the bias towards BFS or DFS.
By combining $\beta$ and the kernel function $\mathcal{K(\cdot)}$ to
compute the transition probabilities, we are able to effectively find
historical neighborhoods where nodes not only have high affinity with
$x$ or $y$ based on the static network structure, but also have high
relevance to the formation of $(x,y)$.
Moreover, allowing duplicate visits in the temporal random walks can
effectively mitigate any potential sparsity issues in the temporal
random walk space.
Specifying the number of walk steps as a walk length is a standard
procedure applied in random walk algorithms, and each random walk
will not stop until that number of steps is reached.
In our temporal random walk algorithm, we only visit relevant nodes,
which are defined using the constraints shown in
Definition~\ref{relevant}.
If backtracking is not allowed and no unvisited neighbor is
``relevant'', the temporal random walk will early terminate before
the pre-defined number of walk steps can be reached.

\subsection{Node-level Attention and Aggregation}
\begin{algorithm}[!t]
 \caption{Forward propagation of the aggregation process}\label{HNA}\label{forward}
   \SetKwInOut{Input}{Input}
   \SetKwInOut{Output}{Output}
  \Input{Graph $G=(V,E)$, the target edge $(x,y)$ and target node $x$, node embeddings \{$\bm{e}_v, \forall v \in \mathcal{V}$ \}, a weight matrix $W$, the number of walks $k$, walk length $\ell$. }
  \Output{Aggregated embedding $\bm{z}_x$ of $x$ after aggregation.}
  \For{ $i = 1$ to $k$}{
        $S_i \leftarrow $ nodes in random walk $\textbf{\textsf{r}}_i$ of length $\ell$ \;
        Compute $\alpha_{v,x}$ for  \{$\bm{e}_v, \forall v \in S_i$ \} \; 
        $\bm{h}_{\textbf{\textsf{r}}_i} =   \text{Relu}\big(\text{BN}\big(\text{LSTM}\big(  \{\alpha_{v,x}\bm{e}_v, \forall v \in S_i \} \big)\big)\big)$ \;}
 Compute $\beta_{\textbf{\textsf{r}}_i,x}$ for \{$\textbf{\textsf{r}}_i, \forall i \in \{1, 2, ..., k \}$ \} \;
   $\bm{{H}} \leftarrow \text{BN}$(LSTM(\{$\beta_{\textbf{\textsf{r}}_i,x}\bm{h}_{\textbf{\textsf{r}}_i}, \forall i \in \{1, 2, ..., k \}$ \})) \;
$ \bm{z}_x  \leftarrow W \cdot [\bm{H} || \bm{e}_x]$  \;
Return $\bm{z}_x \leftarrow \bm{z}_x /  \| \bm{z}_x\|_2$ \;
\end{algorithm}

During first level aggregation, we combine information from nodes
appearing in the same random walks instead of processing all nodes in
a historical neighborhood deterministically.
This strategy has two benefits: (1) it reduces the number of nodes
being aggregated, while still improving the effectiveness (only the
most relevant nodes are covered), and can lead to faster convergence;
(2) each random walk can be roughly regarded as a sequence of
chronological events (e.g., edge formations) which can be leveraged
to capture more meaningful information (i.e., the sequential
interactions/relationships between nodes).
In a manner commonly used for sequential data in the NLP community,
each random walk can be viewed as a `sentence', and a stacked LSTM
(multi-layer LSTM) can then be applied as the aggregator.
Below is a sketch of the workflow of node-level aggregation: 

$$
\bm{h} =
\text{Relu}\big(\text{BN}\big(\text{AGGREGATE\big(Input\textsubscript{1}\big)}\big)\big)
$$

\noindent where (1) Input\textsubscript{1} = \{$\bm{e}_v, \forall v
\in S_i$ \} refers to the input at the first level (i.e., node-level)
where $S_i$ refers to the ordered set of nodes co-occurring in the
$i$-th temporal random walk $\textbf{\textsf{r}}_i$ and $\bm{e}_v$
refers to the learned embedding of node $v$; (2) AGGREGATE refers to
the stacked LSTM aggregator function; (3) $\text{BN}$ refers to the
batch normalization~\cite{batchnorm} and we apply it and mini-batch
(i.e., 512) stochastic gradient descent to our method in experiments;
(4) $\text{Relu}(x)=max(0,x)$ is an element-wise non-linear
activation; (5) $\bm{h}_{\textbf{\textsf{r}}_i }$ refers to the
representation of the $i$-th random walk $\textbf{\textsf{r}}_i$
which captures and summarizes useful information from nodes appearing
in the random walk.

During the aggregation, nodes are treated differently based on their
relevance to the edge formation.
There are three factors determining the relevance of a node $v$ to
the formation of the target edge: (1) the timing information of
(in)direct interactions between $v$ and the target node; (2) the
interaction frequency; (3) the relationship between $v$ and the
target node in the embedding space.

Suppose we have a random walk $\textbf{\textsf{r}}$ initiated from
the target node $x$.
Inspired by the recent attention based mechanisms for neural machine
translation~\cite{attention}, if node $v$ appears in
$\textbf{\textsf{r}}$, we can define an attention coefficient
$\alpha_{v,x}$ of $v$ applied to the aggregation process using
Softmax: 

\begin{equation}\label{coefficient1}
\alpha_{v,x} = \frac{ \exp(-  \frac{1}{ \sum\limits_{ (u,v) \; \text{in} \;  \textbf{\textsf{r}} } t_{(u,v)}  }  \| \bm{e}_x-\bm{e}_v \|_2^2)  }{ \sum\limits_{v' \; in \;  \textbf{\textsf{r}}} \exp(-  \frac{1}{ \sum\limits_{ (u,v') \; \text{in} \;  \textbf{\textsf{r}} } t_{(u,v')}  }  \| \bm{e}_x-\bm{e}_{v'} \|_2^2)  }
\end{equation}

As shown in Equation~\ref{coefficient1}, in the random walk
$\textbf{\textsf{r}}$ initiated from $x$, the attention coefficient
$\alpha_{v,x}$ of $v$ has a positive correlation with the temporal 
information of interactions (i.e., $t_{(u,v)}$) and the interaction
frequency (i.e., $\sum_{ (u,v) \; \text{in} \; \textbf{\textsf{r}}
}$), and it has a negative correlation with the similarity between
$x$ and $v$ in the embedding space.

Therefore, we can update the Input\textsubscript{1} with node
embeddings weighted by the computed attention coefficients such that 
$$ 
\text{Input} \textsubscript{1} = \{ \alpha_{v,x} \bm{e}_v, \forall v \in S_i \}
$$

\subsection{Walk-level Attention and Aggregation}
In second level aggregation, the goal is to combine information from
each representation $\bm{h}$ learned during the first stage of random
walks to learn a new embedding $\bm{e}_x$ that contains information
from both the target node $x$ and all of the surrounding relevant
nodes.
So, second stage walk-level aggregation can now be formalized as: 

$$
\bm{H}=
\text{BN}\big(\text{AGGREGATE\big(Input\textsubscript{2}\big)}\big)
$$
$$
\bm{e}_x =W \cdot [\bm{H} || \bm{e}_x]
$$

\noindent where (1) Input\textsubscript{2}=
\{$\bm{h}_{\textbf{\textsf{r}}_i}, \forall i \in \{1, 2, ..., k \}$
\} refers to the input at the second level where
$\bm{h}_{\textbf{\textsf{r}}_i}$ is the representation of the $i$-th
random walk $\textbf{\textsf{r}}_i$; (2) $\bm{H}$ is the
representation of the historical neighborhood which captures and
summarizes information from representations of temporal random walks;
(3) $W$ is a trainable weight matrix and $[\cdot || \cdot]$ refers to
the concatenation operation which can be viewed as a simple form of a
`skip connection'~\cite{he2016identity} between different levels in
the EHNA model.

Similar to node-level aggregation, random walks should be treated
differently based on their relevance to the formation of the target
edge $(x,y)$.
We consider three factors when determining the relevance of a random
walk $\textbf{\textsf{r}}$ initiated from $x$ or $y$: (1) the average
timing information of (in)direct interactions between nodes appearing
in $r$ and the target node; (2) the average interaction frequency of
nodes appearing in $\textbf{\textsf{r}}$; (3) the similarity between
the random walk and the target node in the embedding space.

Thus, given the set $\textbf{\textsf{R}}$ of all random walks and a
target node $x$, we define the attention coefficient $\beta_{
\textbf{\textsf{r}},x}$ of a random walk $\textbf{\textsf{r}}$
applied to the aggregation process as: 

\begin{equation}\label{coefficient2}
\beta_{\textbf{\textsf{r}},x} = \frac{ \exp(- \frac{1}{|\textbf{\textsf{r}}|} (\sum\limits_{v \; in \; \textbf{\textsf{r}}} \frac{1}{ \sum\limits_{ (u,v) \; \text{in} \; \textbf{\textsf{r}} } t_{(u,v)}  })  \| \bm{e}_x-\bm{h}_{\textbf{\textsf{r}}} \|_2^2)  }{ \sum\limits_{\textbf{\textsf{r}}' \; in \; \textbf{\textsf{R}}} \exp(-  \frac{1}{|\textbf{\textsf{r}}'|}  (\sum\limits_{v \; in \; \textbf{\textsf{r}}'}  \frac{1}{ \sum\limits_{ (u,v') \; \text{in} \; \textbf{\textsf{r}} } t_{(u,v')}  })  \| \bm{e}_x-\bm{h}_{\textbf{\textsf{r}}'} \|_2^2)  }
\end{equation}

\noindent where $|\textbf{\textsf{r}}|$ refers to the number of nodes
appearing in the random walk $\textbf{\textsf{r}}$.
Therefore, we can update Input\textsubscript{2} with representations
of random walks weighted by the computed attention coefficients such
that 

$$ 
\text{Input} \textsubscript{2} =\{ \beta_{\textbf{\textsf{r}}_i,x}\bm{h}_{\textbf{\textsf{r}}_i }, \forall i \in \{1, 2, ..., k \} \}
$$

Algorithm~\ref{forward} describes the forward propagation of the
aggregation process for a model that has already been trained using a
fixed set of parameters.
In particular, lines 1-4 correspond to the node-level attention and
aggregation process, and lines 5-8 refer to the walk-level attention
and aggregation process.

\subsection{Optimization} \label{sec_objective}
To analyze the formation of each edge $(x,y)$, we have described how
our model generates \textsl{aggregated embeddings} $\bm{z}_x$ and
$\bm{z}_y$ for target nodes $x$ or $y$ using information aggregation
of historical neighborhoods.
Then, we describe our objective function and how it is optimized.

Unlike previous work on network
embeddings~\cite{perozzi2014deepwalk,grover2016node2vec} which apply 
a distance independent learning measure (e.g., dot product) to
compute the proximity between nodes, we have adopted Euclidean
distance as it satisfies the triangle inequality, which has been shown
to be an important property when generalizing learned
metrics~\cite{hsieh2017collaborative,chen2018pme}.
Based on the triangle inequality, the distance of any pair cannot be
greater than the sum of the remaining two pairwise distances.
Thus, by leveraging Euclidean distance, we can naturally preserve the
first-order (the pairwise proximity between nodes) and second-order
proximity (the pairwise proximity between node neighborhood
structures)~\cite{wang2016structural}.

Our goal is to keep nodes with links close to each other and nodes
without links far apart in the embedding space.
More specifically, with the set $\mathcal{S}=\{(\bm{z}_x,\bm{z}_y),
\forall (x,y) \in \mathcal{E} \}$ of pairwise aggregated embeddings
for all edges, we define the following margin-based objective loss 
function:
\begin{equation}\label{objective1}
\mathcal{L}= \sum_{(\bm{z}_x,\bm{z}_y) \in \mathcal{S}} \sum_{(\bm{z}_x,\bm{z}_u) \notin \mathcal{S}, u \in \mathcal{V} } [m+ \| \bm{z}_x-\bm{z}_y \|_2^2 - \| \bm{z}_x-\bm{z}_u \|^2_2 ]_+
\end{equation}

\noindent where $m$ is the safety margin size and $[\cdot]_+ =
max(\cdot,0)$ is the hinge loss.
Note that, for nodes in the set $\{u | \forall u, (\bm{z}_x,\bm{z}_u)
\notin \mathcal{S}\}$, we cannot generate their aggregated embeddings
since we cannot identify historical neighborhoods for them without
additional edge information.
For such nodes, we aggregate information from neighborhoods by
randomly sampling node connections two hops away, in a manner similar
to {\method{GraphSAGE}}~\cite{hamilton2017inductive} whose
aggregation strategy was proven to be an instance of the
Weisfeiler-Lehman isomorphism~\cite{shervashidze2011weisfeiler}, and
is capable of approximating clustering coefficients at arbitrary
degrees of precision.

However, it is computationally expensive to minimize
Equation~\ref{objective1} since there is a large number of node pairs
without links.
Thankfully, our solution can utilize negative sampling
techniques~\cite{mikolov2013distributed} to mitigate this problem.
To apply the technique, the most likely negative samples are gathered
based on node degree, in a manner similar to sampling negative words
based on frequency in the NLP community.
Specifically, we sample negative nodes based on the widely adopted
noise distribution~\cite{ahmed2013distributed} $P_n(v) \sim
d_v^{0.75}$, where $d_v$ is the degree of node $v$.
Our objective function can now be rewritten as:
\begin{multline}\label{objective2}
\mathcal{L} = \sum_{(\bm{z}_x,\bm{z}_y) \in \mathcal{S}}\\
 (\sum_{q=1}^{\mathcal{Q}} \mathbb{E}_{v_q} \sim P_n(v) [m+ \| \bm{z}_x-\bm{z}_y \|_2^2 -  \| \bm{z}_x-\bm{z}_{v_q} \|^2_2 ]_+ )
\end{multline}

\noindent where $\mathcal{Q}$ is the number of negative samples.

In Equation~\ref{objective2}, we only generate negative samples in 
one direction.
However, it may not be sufficient to apply this negative sampling
strategy in certain networks, especially heterogeneous networks.
For example, in the e-commerce network Tmall which contains the
`buyer-item' relationships, if we only sample negative samples
from the item side, we cannot effectively learn embeddings for
purchaser nodes.
However, we can somewhat mitigate this limitation by applying
bidirectional negative sampling~\cite{yin2017sptf} to
produce the objective function:
\begin{multline}\label{final_objective}
\mathcal{L} = \sum_{(\bm{z}_x,\bm{z}_y) \in \mathcal{S}}\\
 (\sum_{q=1}^{\mathcal{Q}} \mathbb{E}_{v_q} \sim P_n(v) [m+ \| \bm{z}_x-\bm{z}_y \|_2^2 - \| \bm{z}_x-\bm{z}_{v_q} \|^2_2 ]_+  \\
+  \sum_{q=1}^{\mathcal{Q}} \mathbb{E}_{v_q} \sim P_n(v) [m+ \| \bm{z}_x-\bm{z}_y \|_2^2 - \| \bm{z}_y-\bm{z}_{v_q} \|^2_2]_+)
\end{multline}

Note that the aggregated embedding $\bm{z}$ is a temporary variable
which drives the updates of embedding $\bm{e}$ of nearby nodes and
other parameters in the framework during the back propagation.
After the learning procedure, we will apply one additional
aggregation process for each node with its most recent edge, and the
aggregated embedding $\bm{z}_x$ generated will be used as the final
embedding of $x$ ($\bm{e}_x=\bm{z}_x$).

%% file: tex/experiment.tex
\section{Experiments}\label{experiment}
In this section, we conduct experiments using several real-world
datasets to compare our method with state-of-the-art network
embedding approaches on two tasks: network reconstruction and link
prediction.
The empirical results clearly show the value of capturing temporal
information during the learning process.

\input{./tex/dataset.tex}
\input{./tex/network.tex}

\subsection{Datasets}
Since the focus of this work is on temporal network representation
learning, we conduct extensive experiments on four real-world
datasets containing temporal information.
They include one e-commerce network, one review network, one
co-author network and one social network.
The statistics of each dataset are summarized in Table~\ref{dataset}
and the detailed descriptions are listed as follows.

\begin{itemize}
\item DBLP 
is a co-author network where each edge represents a co-authorship
between two authors (nodes).
We derive a subset of the co-author network which contains
co-authorship between researchers from 1955 to 2017.

\item Digg
is a social network where each edge represents the friendship between
two users (nodes) and each edge is annotated with a timestamp,
indicating when this friendship is formed.
The time span of the network is from 2004 to 2009.

\item Tmall \cite{zuo2018embedding} is extracted from the sales data
of the ``Double 11'' shopping event in 2014 at
Tmall.com\footnote{https://tianchi.aliyun.com/datalab/dataSet.htm?id=5}.
Each node refers to either one user or one item and
each edge refers to one purchase with a purchase date.

\item Yelp \cite{zuo2018embedding} is extracted from the
Yelp\footnote{https://www.yelp.com} Challenge Dataset.
In this network, each node either refers to one user or one business,
and each comment is represented as an edge connection between users.
\end{itemize}

\subsection{Baseline Algorithms}
We use the following four methods as the baselines. 
\begin{itemize}
\item \method{HTNE} \cite{zuo2018embedding} adopted the Hawkes
process~\cite{hawkes1971spectra} to model the neighborhood formation
sequences where the current neighbor formation can be influenced with
higher intensity by more recent events. 

\item \method{Node2Vec} \cite{grover2016node2vec} first sampled
random walks with some parameters controlling
inward/outward explorations (e.g., DFS and BFS) to cover both micro
and macro views of neighborhoods.
Then random walks are sampled to generate sequences of nodes which
can be used as the input to a skip-gram model to learn the final
representations.

\item \method{CTDNE}~\cite{nguyen2018continuous} extended the idea of
\method{Node2Vec}~\cite{grover2016node2vec} with the skip-gram model
by adding time constraints on the random walks and require nodes to
co-occur in the same time-constrained random walk in order to produce
similar embeddings for related nodes.

\item \method{LINE} \cite{tang2015line} optimized node
representations by preserving the first-order and second-order
proximity of a network.
As recommended by the authors, we concatenate the representations
that preserve these two proximities respectively and use the
concatenated embeddings for our experiments.
\end{itemize}
\subsection{Parameter Settings}
For our method, we set the mini-batch size, the safety margin size
and the number of layers in stacked LSTM to be 512, 5 and 2
respectively.
For parameters $p$ and $q$ in the time-aware random walks and the
learning rate $r$, we use grid search over $p,q \in \{0.25, 0.50, 1,
2, 4\}$ and $r \in \{2 \times 10^{-5}, 2 \times 10^{-6}, 2 \times
10^{-7}\}$ respectively.
We set the number of walks $k=10$ and the walk length $l=10$ by
default.
Following previous 
work~\cite{perozzi2014deepwalk,grover2016node2vec}, we set the
number of negative samples to 5 for all methods and set $k=10$ and
$l=80$ for Node2Vec.
For CTDNE, we use the uniform sampling for initial edge selections
and node selections in the random walks and set the window count to
the same value used by Node2Vec.
The embedding size is fixed to 128 for all methods.
For all other baseline parameters, the recommended were used.

\subsection{Network Reconstruction}
This first task we evaluate is network reconstruction, which is one
of the primary problems originally explored for network embedding
\cite{wang2016structural}.
In this task, we train node embeddings and rank pairs of nodes using 
dot product similarity (one of the most common generic similarity
metrics~\cite{cui2018}) sorted in descending order.
If a pair of nodes is ranked higher based on the similarity metric,
an edge formed by this pair of nodes is more likely to exist in the
original network.
The evaluation metric $Precision@\mathcal{P}$~\cite{cui2018} is
defined as below: $Precision@\mathcal{P}= \frac{1}{\mathcal{P}}
\sum_{i=1}^{\mathcal{P}}\mu_i$.
%
%
Here, $\mathcal{P}$ is the number of evaluated pairs.
$\mu_i =1$ means that the edge formed by the $i-th$ reconstructed
pair of nodes exists in the original network; otherwise, $\mu_i =0$.
Since it is computationally expensive to process all possible
$\frac{|\mathcal{V}|(|\mathcal{V}|-1)}{2}$ pairs of nodes, especially
for large networks, we conduct evaluations on ten thousand randomly
sampled nodes and repeat this process ten times and report the
average $Precision@\mathcal{P}$.

Figure~\ref{reconstruct} compares the precision scores achieved by
different methods on the four datasets.
Our method \method{EHNA} outperforms all baselines across all
datasets.
More specifically, \method{EHNA} consistently outperforms
\method{HTNE} and \method{CTDNE} when the number $\mathcal{P}$ of
reconstruction pairs varies from $1 \times 10^2$ to $1 \times 10^6$,
outperforms \method{Node2Vec} when $\mathcal{P}$ varies from 500 to
$1 \times 10^6$, and outperforms \method{LINE} 
when $\mathcal{P}$ varies from $1 \times 10^4$ to $1 \times 10^6$.
Note that the difference w.r.t.
the performance of some methods can be very small when the number of
pairs of nodes is small, thus it is very hard to visually distinguish
these methods.
However, the difference w.r.t.
the performance between \method{EHNA} and \method{HTNE},
\method{Node2Vec} or \method{CTDNE} is notable on all datasets under
different cases.

All methods converge to similar performance when $\mathcal{P}$ is
sufficiently large since most of the true edges which were previously ranked
have already been recovered correctly.
The above results show that our method can effectively capture one of the most
important aspects of network structure when leveraging the available
temporal information.

\input{./tex/operator.tex}
\input{./tex/linkt_digg.tex}

\input{./tex/linkt_yelp.tex}

\input{./tex/linkt_tmall.tex}

\input{./tex/linkt_dblp.tex}

\subsection{Link Prediction}\label{sec_link}
The link prediction task aims to predict which pairs of nodes are
likely to form edges, which is an important application of network
embedding.
In this task, we focus on future link prediction.
To simulate this problem, we remove 20\% of the most recent edges in 
a graph, and use them for prediction.
The rest of the network is used to train the model.
After the training process, we use the removed edges as the positive
examples and generate an equal number of negative examples by 
randomly sampling node pairs from the network where there are no
current edge connections.
After generating the positive and negative examples, we generate edge
representations using the learned node embeddings and split these
edge representations into a training dataset and test dataset which
then are used to train the same logistic regression classifier with
the LIBLINEAR package~\cite{fan2008liblinear} to ensure that all
embeddings were compared on an equal footing.

We generate edge representations with learned node embeddings based
on several commonly used binary operators.
More specifically, given a binary operator~$\circ$, as well as two
embeddings $\bm{e}_x$ and $\bm{e}_y$ of node $x$ and $y$
respectively, we generate the representation $f(x,y)$ of the edge
$(x,y)$ such that $f : V \times V \rightarrow \mathbb{R}^d$.
Here, $d=128$ is the representation size for edge $(x,y)$.
Following previous work~\cite{grover2016node2vec}, we use four
different operators (Table~\ref{operator}) to generate the final edge
representations.
These four operators can be interpreted as four different indications
of the relationships between the representations of nodes with links
and nodes without links.
For example, the mean operator may indicate that the averaged
vectors of pairs of nodes with links should be located in similar
positions in the embedding space.
Similarly, the Weighted-L1 and L2 operators may indicate that the
distance between nodes with links in the embedding space should be
smaller than the one between nodes without links.

If the node embeddings learned by a method preserve the
property or relationship indicated by a specific operator, then a
classifier can easily distinguish positive edge representations
(i.e., edges actually exist in the graph) from negative edge
representations (i.e., edges which do not exist in the graph)
generated by that operator.
In the link prediction task, many existing studies have only adopted
one operator for the final evaluation.
Since the choice of operator may be domain specific, an effective
embedding method should be robust to the choice of operators.
Thus, we report the performance results under all operators defined
in Table~\ref{operator}.
Some operators clearly work better than others, but we are unaware
of any systematic and sensible evaluation of combining operators or how to chose
the best one for network embedding problems such as link prediction.
We leave this exploration to further work.

We randomly select 50\% of examples as the training dataset and the
rest of examples as the test dataset and train the logistic
regression classifier with the LIBLINEAR
package~\cite{fan2008liblinear}.
We repeat this process ten times and report the average performance.
Tables~\ref{link_digg}-\ref{link_dblp} compare the
performance of all methods with different combining operators
on each dataset.
Our method \method{EHNA} significantly outperforms all other baselines
in most cases.
In the cases where \method{EHNA} is not the best performer, the
performance of \method{EHNA} is still quite competitive, achieving
the second best result. 
Note that we compute the error reduction by comparing \method{EHNA}
with the best performer in baselines and the performance of baselines
varies notably across different datasets.
Thus, the performance gains of \method{EHNA} is more
significant when making pairwise comparisons with 
certain other baselines.

\input{./tex/parameter.tex}

\subsection{Ablation Study}
Table~\ref{ablation} compares the performance of different variants
of \method{EHNA} in the link prediction task.
Due to space limitations, we only show the performance using the
Weighted-L2 operator.
\method{EHNA-NA} refers to the variant without attention mechanisms,
\method{EHNA-RW} refers to the variant which employs traditional
random walks without an attention mechanism and \method{EHNA-SL} refers
to the variant which only uses a single-layered LSTM without the
two-level aggregation strategy.
As we can see in Table~\ref{ablation}, each of our proposed techniques (i.e.,
temporal random walks, the two-level aggregation and attention
models) contributes to the effectiveness of our learned
representations, and affect overall performance in downstream tasks.

\input{./tex/ablation.tex}
\input{./tex/efficiency.tex}
\subsection{Efficiency Study}
In this section, we compare the efficiency of different methods.
Note that it is hard to quantify the exact running time of
learning-related algorithms as there are multiple factors (e.g.,
initial learning rates) that can impact convergence
(e.g., the number of epochs) and we did not over tune hyperparameters
for any method.
Instead, we explore the comparative scalability of the methods
by reporting the average running time in each epoch as
shown in Table~\ref{efficiency} where \method{Node2Vec\_10},
\method{CTDNE\_10} and \method{Line\_10} refer to the multi-threaded
versions (i.e., 10 threads in our experiments) of the corresponding
baselines.
When compared to the baselines, our method \method{EHNA} is scalable
and efficiency may be further improved with multi-threading.
It is worth noting that \method{Line} achieved similar performance
across all datasets as its efficiency depends primarily on the number
of sampled edges which is the same in all experiments.

\subsection{Parameter Analysis}
In this section, we examine how the different choices of parameters
affect the performance of \method{EHNA} on the Yelp dataset with 50\%
as the training ratio.
Excluding the parameter being evaluated, all parameters use their
recommended default as discussed above.

\smallskip
\noindent \textbf{The safety margin size.}
Recall that our objective function (Equation~\ref{final_objective})
aims to keep nodes with links close to each other and nodes without
links far apart.
The safety margin size $m$ controls how close nodes with links can be
and how far nodes without links can be to each other.
Thus, increasing the value of $m$ tends to lead to more accurate
results.
We show how $m$ influences \method{EHNA} in Figure~\ref{margin}.
The performance improves notably when $m$ ranges from 1 to 4 and
converges when $m=5$.

\smallskip
\noindent \textbf{The walk length of temporal random walks.}
The walk length is crucial to the identification of relevant nodes in
historical neighborhoods as it potentially influences the number of 
relevant nodes and specifies the upper bound of how `far' a node can
be from the target node.
As shown in Figure~\ref{walk}, the performance improves significantly
when $l$ ranges from 1 to 10, followed by subtle improvements when
$l$ increases from 10 to 15.
Then, the performance of \method{EHNA} slightly decays when $l$
ranges from 15 to 25.
These observations suggest: (1) nodes `too far away` from the target
node have little impact on the behavior of the target node, and their
inclusion can introduce noise into the aggregation process, resulting
in performance decay; (2) our proposed deep neural network is capable
of capturing and aggregating useful information from relevant nodes
and filtering noisy nodes in historical neighborhoods, but when the
length of random walks become too long, the performance of our method
begins to degrade.

\smallskip
\noindent \textbf{Parameters $p$ and $q$.}
The parameter $p$ controls the likelihood of immediately revisiting
the previous node in a walk and the parameter $q$ controls the bias
towards BFS and DFS.
It is interesting to see how relevant nodes are distributed in the
neighborhood for different choices of $p$ and $q$.
In Figure~\ref{p}, the performance of \method{EHNA} peaks when
$log_2^{p}=-1$.
In Figure~\ref{q}, the performance of \method{EHNA} peaks when
$log_2^{q}=1$.
Recall from Equation~\ref{random} that a small value of $p$ will
cause the walk to backtrack and keep the walk `locally' close to
the starting node (i.e. the target node).
A large $q$ encourages `inward' explorations and BFS behavior.
Thus, the observation from Figure~\ref{p} and Figure~\ref{q}
indicates that the vast majority of relevant nodes reside in nearby
neighborhoods in the Digg dataset.
Thus, by probing the optimal values of $p$ and $q$, we can better
understand the distributions of relevant nodes in different datasets.

\section{Conclusion}\label{conclusion}
In this paper, we studied the problem of temporal network embedding
using historical neighborhood aggregation.
In reality, most of networks are formed through sequential edge
formations in chronological order.
By exploring the most important factors influencing edge formations
in certain classes of networks, we can better model the evolution of
the the entire network, and induce effective feature representations
that capture node influence and interactions.
We first proposed a temporal random walk to identify historical
neighborhoods containing nodes that are most likely to be relevant to
edge formations.
Then we aggregated the information from the target graph in order to
analyze edge formations.
Then, we introduced two optimization strategies, a two-level
aggregation and attention mechanism to enable our algorithm to
capture temporal information in networks.
Extensive experiments on four large scale real-world networks have
demonstrated the benefits of our new approach for the tasks of
network reconstruction and link prediction.

\vspace{.3em}
\textbf{Acknowledgment.} This work was partially supported by ARC DP170102726,
DP180102050, DP170102231, NSFC 61728204, 91646204 and Google. Guoliang Li was partially supported by the 973 Program of China (2015CB358700), NSFC (61632016, 61521002, 61661166012), Huawei, and TAL education.

%% file: tex/dataset.tex
\begin{table}[] 
\centering
\begin{adjustbox}{width=0.40\textwidth}
\begin{tabular}{|c|c|c|}
\hline
\multicolumn{1}{|l|}{\textbf{Datasets}} & \multicolumn{1}{l|}{\textbf{\# nodes}} & \multicolumn{1}{l|}{\textbf{\# temporal edges}} \\ \hline
Digg & 279,630 & 1,731,653 \\ \hline
Yelp & 424,450 & 2,610,143 \\ \hline
Tmall & 577,314 & 4,807,545 \\ \hline
DBLP & 175,000 & 5,881,024 \\ \hline

\end{tabular}
\end{adjustbox}
\caption{The statistics of datasets}\label{dataset}
\vspace{-2em}

\end{table}

%% file: tex/network.tex
\begin{figure*}[!t]

\centering
\includegraphics[width=0.88\textwidth]{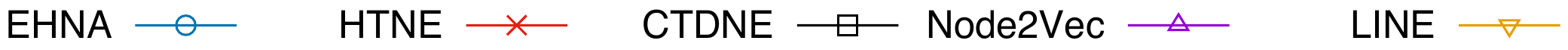}\hspace{-1em}
\hspace{-1em}
\subfloat[Digg]{
\includegraphics[width=0.25\textwidth]{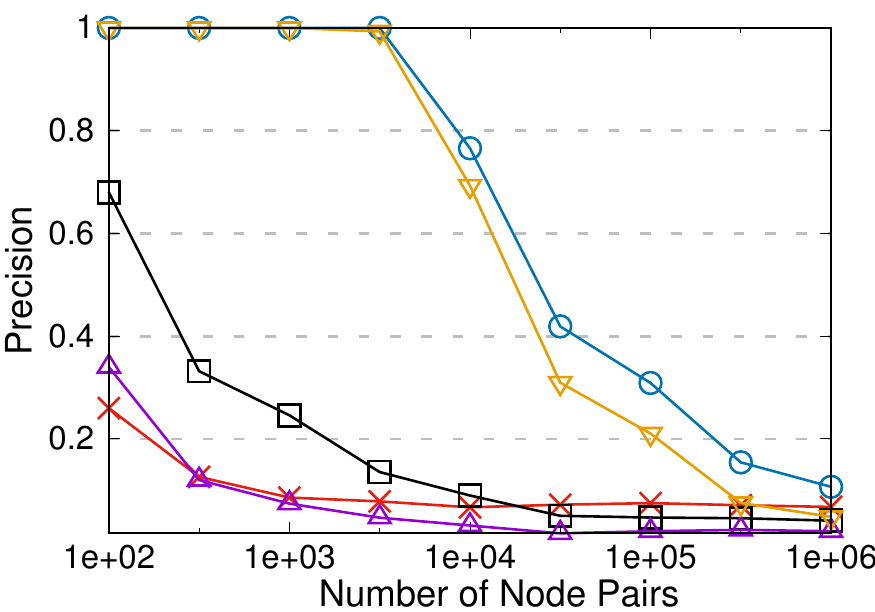}\hspace{-1em}
} 
\subfloat[Yelp]{
\includegraphics[width=.25\textwidth]{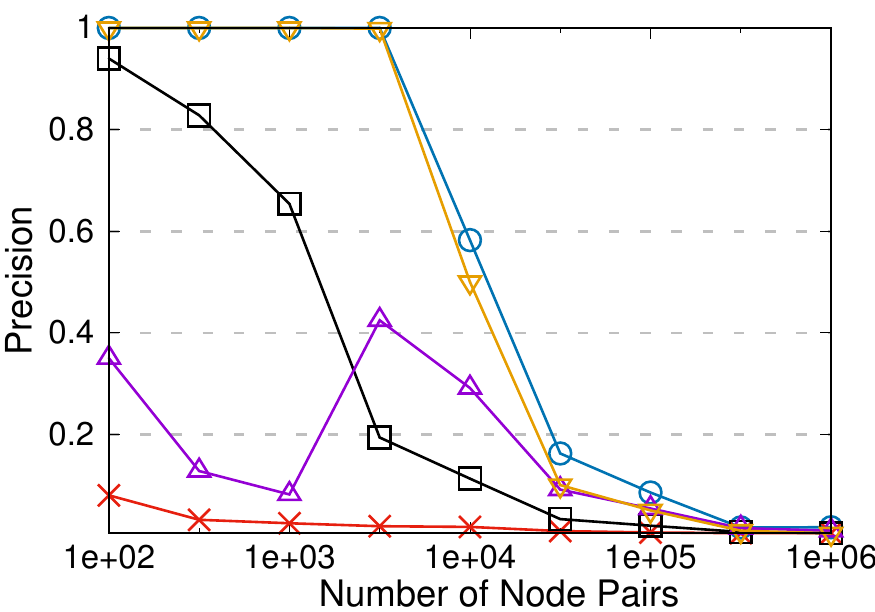}\hspace{-1em}
}
\subfloat[Tmall]{
\includegraphics[width=0.25\textwidth]{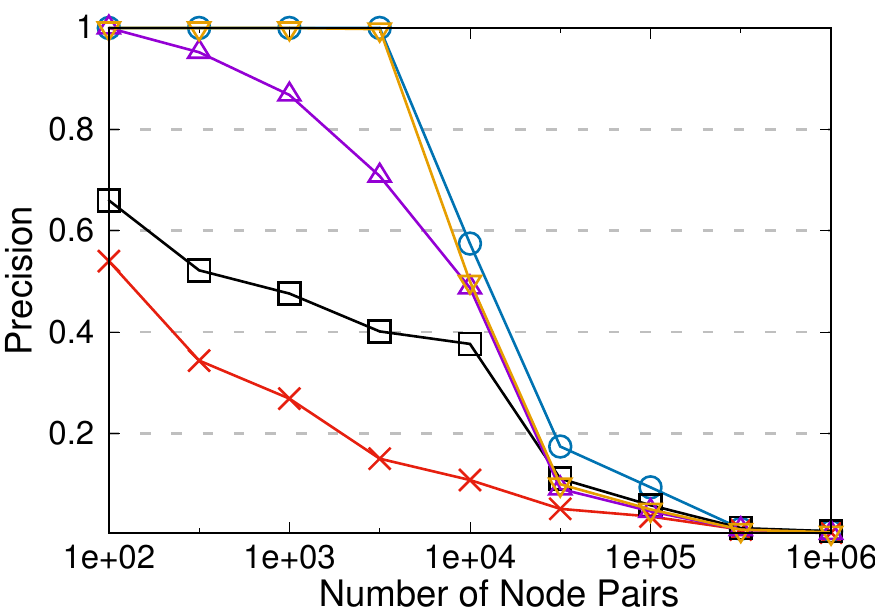}\hspace{-1em}
}
\subfloat[DBLP]{
\includegraphics[width=.25\textwidth]{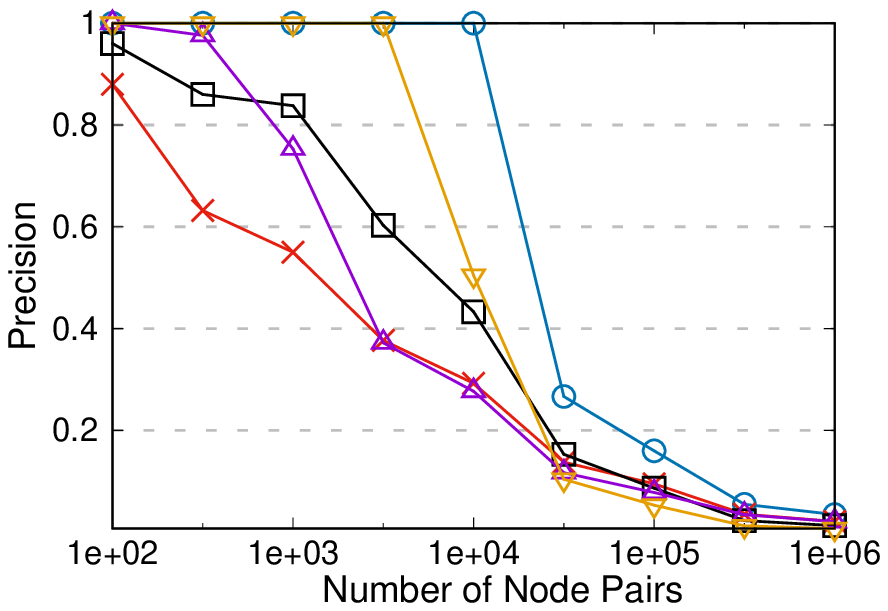}\hspace{-1em}
}
\caption{The precision scores achieved by different methods in the network reconstruction task.}
\label{reconstruct}
\end{figure*}

%% file: tex/operator.tex
\begin{table}[]
\begin{adjustbox}{width=0.48\textwidth}
\centering \begin{tabular}{|c|c|c|}
\hline
\multicolumn{1}{|c|}{\textbf{Operators}} & \multicolumn{1}{c|}{\textbf{Symbol}} & \multicolumn{1}{c|}{\textbf{Definition}} \\ \hline
Mean & $\boxplus$ & $[\bm{e}_x \boxplus \bm{e}_y ]_i = \frac{ \bm{e}_x(i) + \bm{e}_y(i)}{2} $\\ \hline
Hadamard & $\boxdot$ & $ [\bm{e}_x \boxdot \bm{e}_y]_i = \bm{e}_x(i) \ast \bm{e}_y(i)$\\ \hline

Weighted-L1 & $\| \cdot \|_{\widehat{1} }$& $ \| \bm{e}_x \cdot \bm{e}_y \|_{\widehat{1}i} = | \bm{e}_x(i) - \bm{e}_y(i)|$ \\ \hline
Weighted-L2 & $\| \cdot \|_{\widehat{2} }$ & $\| \bm{e}_x \cdot \bm{e}_y \|_{\widehat{2}i} = |  \bm{e}_x(i) - \bm{e}_y(i) |^2$\\ \hline
\end{tabular}
\end{adjustbox}
\caption {Choice of binary operators $\circ$ for edge representations. The definitions correspond to the $i$-th element of~$\bm{e}_{(x,y) }$ and $\bm{e}_x(i)$ refers to the $i$-th element of~$\bm{e}_x$.} \label{operator}
\vspace{-2em}
\end{table}

%% file: tex/linkt_digg.tex
\begin{table*}[]
\begin{adjustbox}{width=2\columnwidth,center}
\tiny
\begin{tabular}{cccccccc}
\hline
Operator & Metric & LINE & Node2Vec & CTDNE & HTNE & EHNA & Error Reduction \\ \hline
\multirow{4}{*}{Mean} & AUC  & \textbf{0.6536} & 0.6322 & 0.6308 & 0.6097 & 0.6404 & -3.8\% \\
 & F1  & 0.6020 & 0.5870 & 0.6149 & 0.5701 & \textbf{0.6634} & 12.6\% \\
 & Precision & 0.6184 & 0.6039 & 0.6683 & 0.5813 & \textbf{0.6881} & 6.0\% \\
 & Recall  & 0.5865 & 0.5711 & 0.5694 & 0.5593 & \textbf{0.6404} & 16.1\% \\ \hline
\multirow{4}{*}{Hadamard} & AUC  & 0.6855 & 0.8680 & 0.9280 & 0.7680 & \textbf{0.9292} & 1.7\% \\
 & F1  & 0.6251 & 0.7969 & 0.8631 & 0.6879 & \textbf{0.8636} & 0.3\% \\
 & Precision  & 0.6370 & 0.8131 & \textbf{0.9132} & 0.7770 & 0.8808 & -37.3\% \\
 & Recall  & 0.6136 & 0.7813 & 0.8182 & 0.6171 & \textbf{0.8469} & 15.8\% \\ \hline
\multirow{4}{*}{Weighted-L1} & AUC & 0.7688 & 0.6788 & \textbf{0.9063} & 0.8237 & 0.9031 & -3.4\% \\
 & F1 & 0.6938 & 0.5843 & \textbf{0.8384} & 0.7481 & 0.8273 & -6.7\% \\
 & Precision  & 0.7085 & 0.6293 & 0.8276 & 0.7458 & \textbf{0.8352} & 4.4\% \\
 & Recall  & 0.6798 & 0.5506 & \textbf{0.8495} & 0.7504 & 0.8196 & -19.9\% \\ \hline
\multirow{4}{*}{Weighted-L2} & AUC  & 0.7737 & 0.6722 & \textbf{0.9057} & 0.8211 & 0.9025 & -3.4\% \\
 & F1  & 0.6999 & 0.5510 & \textbf{0.8296} & 0.7540 & 0.8267 & -1.7\% \\
 & Precision  & 0.7119 & 0.6497 & \textbf{0.8493} & 0.7341 & 0.8092 & -26.6\% \\
 & Recall  & 0.6882 & 0.4783 & 0.8107 & 0.7750 & \textbf{0.8405} & 19.6\% \\ \hline
\end{tabular}
\end{adjustbox}
\caption{Different metric scores achieved by methods with different operators in the link prediction task on Digg.Error Reduction is calculated as $\frac{(1-\text{them})-(1-\text{us})}{(1-\text{them})}$~\cite{abu2018watch}, where ``them'' refers to the best performer from baselines and ``us'' refers to our proposed method \method{EHNA}.}\label{link_digg}
\end{table*}

%% file: tex/linkt_yelp.tex
\begin{table*}[]
\begin{adjustbox}{width=2\columnwidth,center}
\tiny
\begin{tabular}{cccccccc}
\hline
Operator & Metric  & LINE & Node2Vec & CTDNE & HTNE & EHNA & Error Reduction \\ \hline
\multirow{4}{*}{Mean} & AUC  & \textbf{0.7669} & 0.5359 & 0.7187 & 0.5167 & 0.7550 & -5.1\% \\
 & F1 & 0.6968 & 0.5261 & 0.6715 & 0.4942 & \textbf{0.7008} & 1.3\% \\
 & Precision & \textbf{0.7147} & 0.5275 & 0.7079 & 0.5018 & 0.6873 & -9.6\% \\
 & Recall  & 0.6797 & 0.5246 & 0.6387 & 0.4868 & \textbf{0.7184} & 12.1\% \\ \hline
\multirow{4}{*}{Hadamard} & AUC  & 0.5683 & 0.9359 & 0.9564 & 0.9497 & \textbf{0.9775} & 48.4\% \\
 & F1  & 0.5500 & 0.8648 & 0.8944 & 0.8911 & \textbf{0.9296} & 33.3\% \\
 & Precision & 0.5506 & 0.8639 & \textbf{0.9231} & 0.9040 & 0.9207 & -3.1\% \\
 & Recall  & 0.5493 & 0.8657 & 0.8674 & 0.8785 & \textbf{0.9387} & 49.5\% \\ \hline
\multirow{4}{*}{Weighted-L1} & AUC  & 0.7611 & 0.8713 & 0.8380 & 0.9413 & \textbf{0.9506} & 15.8\% \\
 & F1  & 0.6891 & 0.8119 & 0.7542 & 0.8776 & \textbf{0.8951} & 14.3\% \\
 & Precision  & 0.6980 & 0.7931 & 0.7744 & 0.8547 & \textbf{0.8739} & 13.2\% \\
 & Recall  & 0.6803 & 0.8315 & 0.7350 & 0.9016 & \textbf{0.9173} & 16.0\% \\ \hline
\multirow{4}{*}{Weighted-L2} & AUC  & 0.7736 & 0.8723 & 0.8296 & 0.9394 & \textbf{0.9465} & 11.7\% \\
 & F1  & 0.7010 & 0.8180 & 0.7280 & 0.8752 & \textbf{0.8895} & 11.5\% \\
 & Precision & 0.7088 & 0.7877 & 0.7911 & 0.8362 & \textbf{0.8527} & 10.1\% \\
 & Recall & 0.6933 & 0.8508 & 0.6742 & 0.9181 & \textbf{0.9296} & 14.0\% \\ \hline
\end{tabular}
\end{adjustbox}
\caption{Different metric scores achieved by methods with different operators in the link prediction task on Yelp.}\label{link_yelp}
\end{table*}

%% file: tex/linkt_tmall.tex
\begin{table*}[]
\begin{adjustbox}{width=2\columnwidth,center}
\tiny
\begin{tabular}{cccccccc}
\hline
Operator & Metric  & LINE & Node2Vec & CTDNE & HTNE & EHNA & Error Reduction \\ \hline
\multirow{4}{*}{Mean} & AUC  & 0.5198 & 0.5643 & \textbf{0.7948} & 0.5277 & 0.7858 & -4.4\% \\
 & F1  & 0.5126 & 0.5542 & \textbf{0.7366} & 0.5182 & 0.7291 & -2.8\% \\
 & Precision  & 0.5139 & 0.5495 & \textbf{0.7330} & 0.5183 & 0.7100 & -8.6\% \\
 & Recall  & 0.5113 & 0.5589 & 0.7403 & 0.5180 & \textbf{0.7492} & 3.4\% \\ \hline
\multirow{4}{*}{Hadamard} & AUC  & 0.5008 & 0.8890 & 0.8704 & 0.8889 & \textbf{0.9407} & 46.6\% \\
 & F1 & 0.4964 & 0.8142 & 0.7838 & 0.8049 & \textbf{0.8707} & 30.4\% \\
 & Precision  & 0.5000 & \textbf{0.8591} & 0.8415 & 0.8294 & 0.8420 & -12.1\% \\
 & Recall  & 0.4928 & 0.7738 & 0.7336 & 0.7817 & \textbf{0.9013} & 54.8\% \\ \hline
\multirow{4}{*}{Weighted-L1} & AUC  & 0.6078 & 0.8205 & 0.6882 & 0.9278 & \textbf{0.9378} & 13.9\% \\
 & F1  & 0.5719 & 0.7407 & 0.6249 & 0.8518 & \textbf{0.8640} & 8.2\% \\
 & Precision  & 0.5754 & 0.7625 & 0.6412 & \textbf{0.8638} & 0.8617 & -1.5\% \\
 & Recall  & 0.5684 & 0.7201 & 0.6093 & 0.8402 & \textbf{0.8664} & 16.4\% \\ \hline
\multirow{4}{*}{Weighted-L2} & AUC & 0.6157 & 0.8239 & 0.6741 & 0.9296 & \textbf{0.9324} & 4.0\% \\
 & F1  & 0.5774 & 0.7439 & 0.6001 & 0.8542 & \textbf{0.8603} & 4.2\% \\
 & Precision  & 0.5798 & 0.7545 & 0.6563 & 0.8525 & \textbf{0.8617} & 6.2\% \\
 & Recall  & 0.5750 & 0.7336 & 0.5527 & 0.8559 & \textbf{0.8664} & 7.3\% \\ \hline
\end{tabular}
\end{adjustbox}
\caption{Different metric scores achieved by methods with different operators in the link prediction task on Tmall.}\label{link_tmall}
\end{table*}

%% file: tex/linkt_dblp.tex
\begin{table*}[]

\begin{adjustbox}{width=2\columnwidth,center}
\tiny
\begin{tabular}{cccccccc}
\hline
Operator & Metric & LINE & Node2Vec & CTDNE & HTNE & EHNA & Error Reduction \\ \hline
\multirow{4}{*}{Mean} & AUC  & 0.5685 & 0.5438 & 0.5763 & 0.5342 & \textbf{0.7362} & 37.7\% \\
 & F1  & 0.5462 & 0.5258 & 0.5277 & 0.4977 & \textbf{0.6735} & 28.1\% \\
 & Precision & 0.5483 & 0.5285 & 0.5447 & 0.5099 & \textbf{0.6024} & 12.0\% \\
 & Recall  & 0.5442 & 0.5231 & 0.5116 & 0.4861 & \textbf{0.7636} & 48.1\% \\ \hline
\multirow{4}{*}{Hadamard} & AUC  & 0.6726 & 0.8770 & 0.8723 & 0.8829 & \textbf{0.9113} & 24.3\% \\
 & F1 & 0.6256 & 0.8311 & 0.8136 & 0.8239 & \textbf{0.8562} & 14.9\% \\
 & Precision & 0.6296 & 0.8233 & \textbf{0.8519} & 0.8274 & 0.8427 & -6.2\% \\
 & Recall & 0.6218 & 0.8391 & 0.7785 & 0.8204 & \textbf{0.8701} & 19.3\% \\ \hline
\multirow{4}{*}{Weighted-L1} & AUC & 0.7147 & 0.8766 & 0.7084 & 0.8971 & \textbf{0.9341} & 36.0\% \\
 & F1 & 0.6532 & 0.8300 & 0.6731 & 0.8486 & \textbf{0.8857} & 24.5\% \\
 & Precision  & 0.6624 & 0.8384 & 0.6402 & 0.8466 & \textbf{0.8675} & 13.6\% \\
 & Recall & 0.6444 & 0.8217 & 0.7095 & 0.8507 & \textbf{0.9046} & 36.1\% \\ \hline
\multirow{4}{*}{Weighted-L2} & AUC & 0.7144 & 0.8775 & 0.7011 & 0.8983 & \textbf{0.9265} & 27.7\% \\
 & F1  & 0.6544 & 0.8364 & 0.6786 & 0.8567 & \textbf{0.8774} & 14.4\% \\
 & Precision  & 0.6599 & 0.8274 & 0.6226 & 0.8330 & \textbf{0.8561} & 13.8\% \\
 & Recall  & 0.6491 & 0.8456 & 0.7457 & 0.8817 & \textbf{0.8997} & 15.2\% \\ \hline
\end{tabular}
\end{adjustbox}
\caption{Different metric scores achieved by methods with different operators in the link prediction task on DBLP.}\label{link_dblp}
\end{table*}

%% file: tex/parameter.tex
\begin{figure*}[!t]

\centering
\hspace{-2em}
\subfloat[][Varing the margin size]{
\includegraphics[width=0.25\textwidth]{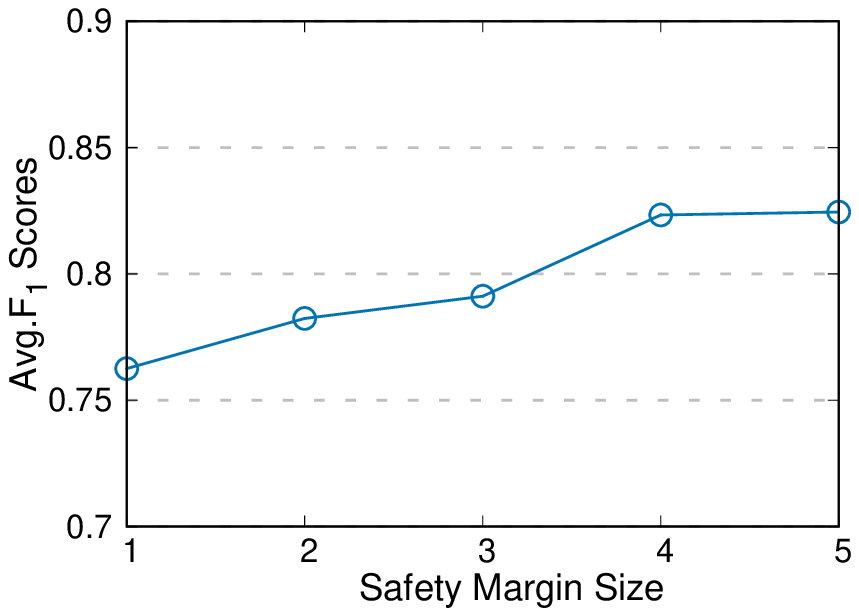}\hspace{-1em}
\label{margin}}
\subfloat[Varing the walk length]{
\includegraphics[width=.25\textwidth]{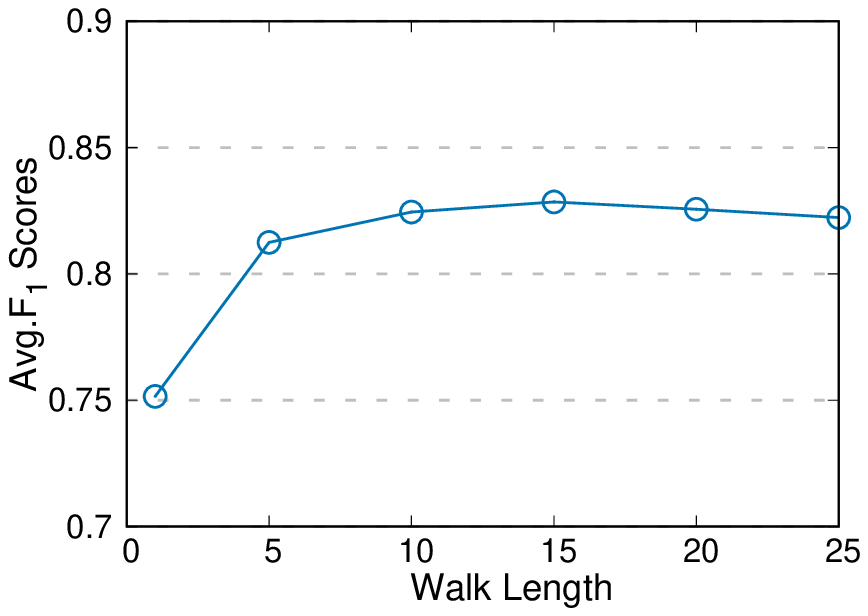}\hspace{-1em}
\label{walk}}
\subfloat[Varing the value of $p$]{
\includegraphics[width=0.25\textwidth]{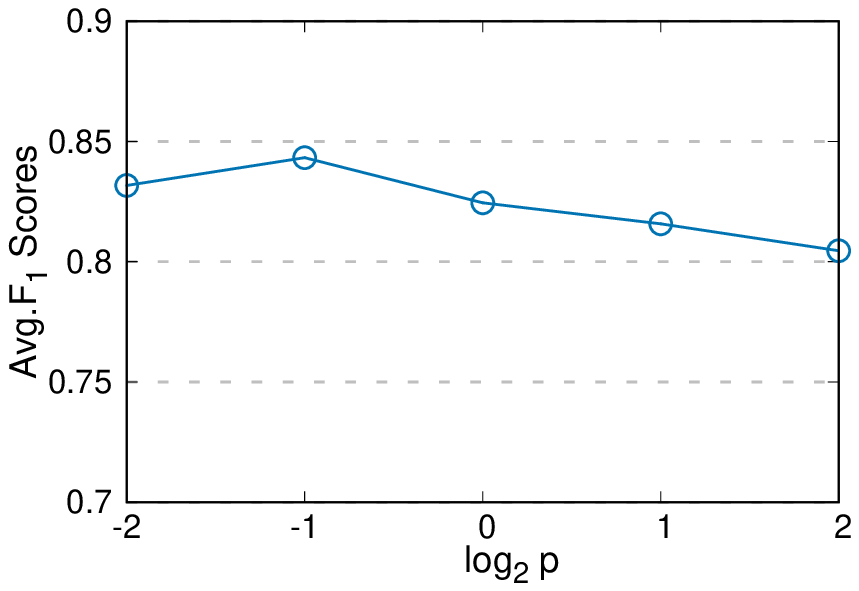}\hspace{-1em}
\label{p}}
\subfloat[Varing the value of $q$]{
\includegraphics[width=0.25\textwidth]{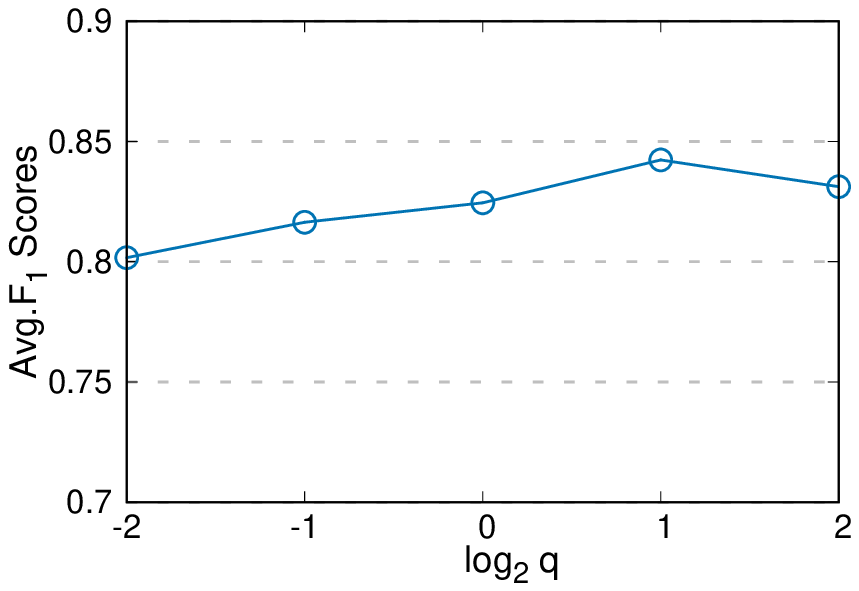}\hspace{-1em}
\label{q}}

\caption{Parameter sensitivity analysis of \method{HTNE} on Yelp.}\label{parameter}

\end{figure*}

%% file: tex/ablation.tex
\begin{table}[]
\centering
\begin{adjustbox}{width=0.48\textwidth}

\begin{tabular}{|c|c|c|c|c|}
\hline
\multirow{2}{*}{\textbf{Method}} & \multicolumn{4}{c|}{\textbf{F1 scores under Weighted-L2}} \\ \cline{2-5} 
 & \textbf{Digg} & \textbf{Yelp} & \textbf{Tmall} & \textbf{DBLP} \\ \hline
EHNA & 0.8267 & 0.8895 & 0.8603 & 0.8774 \\ \hline
EHNA-NA & 0.8131 & 0.8714 & 0.8442 & 0.8685 \\ \hline
EHNA-RW & 0.7837 & 0.8446 & 0.8233 & 0.8327 \\ \hline
EHNA-SL & 0.7254 & 0.7784 & 0.7532 & 0.7231 \\ \hline
\end{tabular}
\end{adjustbox}
\caption{Performance comparisons among different variants of ENHA in the link prediction task.}\label{ablation}
\end{table}

%% file: tex/efficiency.tex
\begin{table}[]
\centering
\begin{adjustbox}{width=0.48\textwidth}
\begin{tabular}{|c|c|c|c|c|}
\hline
\multirow{2}{*}{\textbf{Method}} & \multicolumn{4}{c|}{\textbf{Average running time per epoch (s)}} \\ \cline{2-5} 
 & \textbf{Digg} & \textbf{Yelp} & \textbf{Tmall} & \textbf{DBLP} \\ \hline
Node2Vec & $4.6 \times 10^3$ & $7.1 \times 10^3$ & $1.0 \times 10^4$ & $2.5 \times 10^3$ \\ \hline
Node2Vec\_10 & $4.8 \times 10^2$ & $8.8 \times 10^2$ & $1.2 \times 10^3$ & $3.2 \times 10^2$ \\ \hline
CTDNE & $2.6 \times 10^3$ & $4.2 \times 10^3$ & $9.1 \times 10^3$ & $1.9 \times 10^3$ \\ \hline
CTDNE\_10 & $3.2 \times 10^2$ & $5.4 \times 10^2$ & $1.1 \times 10^3$ & $2.2 \times 10^2$ \\ \hline
Line\_10 & $1.2 \times 10^4$ & $1.2 \times 10^4$ & $1.2 \times 10^4$ & $1.2 \times 10^4$ \\ \hline
HTNE & $3.8 \times 10^1$ & $5.3 \times 10^1$ & $1.1 \times 10^2$ & $1.6 \times 10^2$ \\ \hline
EHNA & $7.8 \times 10^2$ & $1.8 \times 10^3$ & $3.2 \times 10^3$ & $1.7 \times 10^3$ \\ \hline
\end{tabular}
\end{adjustbox}
\caption{Training time comparisons.}\label{efficiency}
\end{table}